# Physics-guided Data Augmentation for Learning the Solution Operator of Linear Differential Equations

**Ye Li, Yiwen Pang, and Bin Shan**

Nanjing University of Aeronautics and Astronautics, Nanjing, 211106, China
yeli20@nuaa.edu.cn

**Abstract:** Neural networks, especially the recent proposed neural operator models, are increasingly being used to find the solution operator of differential equations. Compared to traditional numerical solvers, they are much faster and more efficient in practical applications. However, one critical issue is that training neural operator models require large amount of ground truth data, which usually comes from the slow numerical solvers. In this paper, we propose a physics-guided data augmentation (PGDA) method to improve the accuracy and generalization of neural operator models. Training data is augmented naturally through the physical properties of differential equations such as linearity and translation. We demonstrate the advantage of PGDA on a variety of linear differential equations, showing that PGDA can improve the sample complexity and is robust to distributional shift.

## 1 Introduction

While there is a rapid growth in machine learning in the last 20 years, most researchers concentrated on image recognition and natural language processing tasks [15, 16], and the development of simulating the physical world around us by deep learning is relatively recent [13]. Many physical systems are described by partial differential equations (PDEs). Solving the underlying PDEs is usually analytically intractable. Although there are numerous numerical solvers developed by many mathematicians and physicists in the last half century, classical numerical solvers are expensive in computation and custom designed per PDE class [1].

Recently, there has been much work applying deep learning to learn PDE solvers [2, 30, 10], in which neural operators [20, 18] have drawn the attention of many researchers for their strong ability to learn the mapping between infinite functional spaces. However, current approaches need abundant high-quality data to generalize well, otherwise real-world out-of-distribution test data is difficult to align with training data, see Figure 1 for an illustration. In reality, we have to train a well-generalized neural operator model with limited 'expensive' training data. One potential route is to incorporate symmetries into the forecasting model to improve generalization by Wang et al. [32, 7]. This leads to the so-called equivariant network. However, the design of equivariant layers is a difficult task.

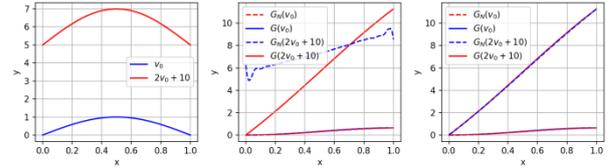

**Figure 1** An illustrative neural operator learning of the antiderivative operator $\mathcal{G}: v(x) \mapsto u(x) = \int_0^x v(\tau)d\tau$ in Section 5.1. (**Left**) Input function $v(x) = v_0(x) = \sin(\pi x)$. (**Middle**) $\mathcal{G}_\mathcal{N}$ without PGDA, then $\mathcal{G}_\mathcal{N}(v_0)$ is well predicted but $\mathcal{G}_\mathcal{N}(2v_0 + 10)$ failed. (**Right**) $\mathcal{G}_\mathcal{N}$ with PGDA, both $\mathcal{G}_\mathcal{N}(v_0)$ and $\mathcal{G}_\mathcal{N}(2v_0 + 10)$ are well predicted.

We propose a simple yet effective data augmentation technique to improve the accuracy and generalization of neural operator models. Training data are augmented naturally through the physical properties of differential equations such as linearity and translation. We name it by physics-guided data augmentation (PGDA), with details in Section 4.2. Our contributions include:

- We study the problem of improving the generalization capability of deep learning models for learning the solution operator of different linear differential equations (of evolutionary/stationary type).

- We design a physics-guided data augmentation (PGDA) method for the training of neural operators. PGDA is based on the physical properties of differential equations such as linearity and translation.

- We demonstrate experimentally that PGDA can improve prediction accuracy under distribution shift up to $1 \sim 4$ orders of magnitude on a SOTA neural operator model.

## 2 Background

Here we present the problem setup of solution operator learning of PDEs. Without loss of generality, we formulate any PDEs as:

$$\begin{cases} \mathcal{L}u(x) = f(x), & x \in \Omega \subset \mathbb{R}^n, \\ \mathcal{B}u(x) = v(x), & x \in \partial\Omega, \end{cases} \quad (1)$$

where $\mathcal{L}$ is the partial differential operator and $\mathcal{B}$ is the boundary condition. Throughout this work we will consider the case where $\mathcal{L}$ and $\mathcal{B}$ are linear operators.

Let $\mathcal{V}$ and $\mathcal{U}$ be the spaces of function $v(x)$ and function $u(x)$, respectively. Then, the mapping $\mathcal{G}: \mathcal{V} \to \mathcal{U}$ with $\mathcal{G}(v)(x) = u(x)$ is referred to the solution operator of the PDE (1). Since the solution operator $\mathcal{G}$ is always nonanalytic, we try to learn an approximate solution operator $\mathcal{G}_\mathcal{N}$ by neural networks and train the network from a training dataset $\mathcal{T} = \{(v^{(1)}, u^{(1)}), (v^{(2)}, u^{(2)}), \cdots, (v^{(N)}, u^{(N)})\}$. Such neural networks $\mathcal{G}_\mathcal{N}$ are named by neural operators, or operator networks.

## 3 Related Work

**Data Augmentation** Although data augmentation techniques are widely used in image applications [28], little work is concerned about the continuous data arising in real-world physical systems such as PDEs. The closest work to ours is Brandstetter et al. [3], who designed a Lie point symmetry data augmentation (LPSDA) technique for neural PDE solvers. Lie point symmetry group of PDE includes time shift, space shift, Galilean boost, scaling, etc, which is very similar to the linearity and transformation in our PGDA method. Our PGDA is also applicable for elliptic PDEs defined on bounded domain with Dirichlet or Neumann boundary conditions, in which symmetries break because open neighborhoods cannot be defined for LPSDA. Luo et al. [23] designed a physics-directed data augmentation technique to reduce target-domain training data sampling complexity in transfer learning. However, a parametric first principle governing the domain shift need to be identified by the user first.

**Equivariant Networks** Developing neural networks that preserve symmetries has been a fundamental task in image recognition [5, 6, 9, 36], but little work on PDE solvers. The pioneering work of Wang et al. [32] incorporated Lie point symmetries into neural network solvers of Navier-Stokes equation and Heat equation. Their models are both theoretically and experimentally robust to distributional shift by symmetry group transformations and enjoy favorable sample complexity. However, the manual design of equivariant networks is not easy. An automatic symmetry discovery and the approximation for imperfectly symmetric dynamics are presented in their subsequent work [7, 33]. A convolutional layer that is equivariant to transformations from any specified Lie group with a surjective exponential map is constructed by Finzi et al. in [9]. Mattheakis et al. [25] embedded even/odd symmetry of a function and energy conservation into neural networks to solve differential equations.

**Physics-informed Deep Learning** Physics-informed deep learning integrating seamlessly data and mathematical physics models (for example, PDEs), has been one of the hotspots in current researches [13, 24]. Physics-informed neural networks (PINNs) proposed by Raissi et al. [27] are one of the outstanding neural PDE solvers in the last few years. Similar work [29, 8, 14, 26] solving PDEs with neural networks has been proposed in the same period. Other techniques include the neural operators [20, 17] presented in Section 4.1, and their variants [34, 35, 18, 12, 22]. There are researches about the so-called neural augmentation where a neural component is added to finite elements [11], multigrid solvers [10], and eikonal solvers [19].

## 4 Methodology

We first briefly introduce two neural operators with promising results in the literatures, namely DeepONet and FNO, then show how the physical properties of PDEs can be used as a data augmentation technique for neural operators.

### 4.1 Neural Operator

**DeepONet** The first neural operator, Deep Operator Network (DeepONet), was published in 2019 by Lu et al. [20] with subsequent theoretical and computational extensions in 2021 [21], and its original architecture was based on the universal approximation theorem of Chen & Chen [4]. Unlike function regression with finite-dimensional inputs and outputs, DeepONet aims to map infinite-dimensional functions (inputs) to infinite-dimensional functions (outputs). More importantly, DeepONet allows for solving PDEs with different boundary and initial conditions without the need for retraining the neural network.

A DeepONet has two sub-networks: a "trunk" network and a "branch" network. The trunk net takes the coordinates $x$ as the input, and a $p$-dimensional vector $[t_1(x), t_2(x), \cdots, t_p(x)]$ as the output. The branch net takes the discretized function $\mathbf{v} = [v(x_1), v(x_2), \cdots, v(x_m)]$ as the input, and the same $p$-dimensional vector $[b_1(\mathbf{v}), b_2(\mathbf{v}), \cdots, b_p(\mathbf{v})]$ as the output. The general output of the DeepONet takes the form:

$$\mathcal{G}_\mathcal{N}(\mathbf{v})(x) = \sum_{k=1}^{p} b_k(\mathbf{v}) t_k(x) + b_0 \quad (2)$$

where $b_0 \in \mathbb{R}$ is a bias. By sampling collocation points $\{y_1, y_2, \cdots, y_M\}$ on the solution function $u(x)$, we get the discrete loss function:

$$L = \frac{1}{NM} \sum_{i=1}^{N} \sum_{j=1}^{M} w_{ij} |u^{(i)}(y_j) - \mathcal{G}_\mathcal{N}(\mathbf{v}^{(i)})(y_j)|^2 \quad (3)$$

where $(\mathbf{v}^{(i)}, y_j)$ are the $N \times M$ different pairs of branch and trunk inputs, $u^{(i)}(y_j)$ is the corresponding labelled outputs, and $w_{ij}$ is the associated weight.

**FNO** The second neural operator, named Fourier Neural Operator (FNO), was published in 2020 by Li et al. [17], and it is based on parameterizing the integral kernel in the Fourier space. The feature of FNO is that the main network parameters are defined and learned in the Fourier space rather than the physical space. One difference between DeepONet and FNO is that FNO discretizes both the input function $v(x)$ and the output function $u(x)$ in the same equispaced mesh.

First, the input function value $v(x)$ is lifted to a higher $d_v$ dimensional representation $z_0(x)$ by

$$z_0(x) = P(v(x)), \quad (4)$$

where $P$ is usually parameterized by a shallow fully connected neural network. Then following an iterative neural operator architecture $z_0 \mapsto z_1 \mapsto \cdots \mapsto z_L$ with

$$z_{t+1}(x) = \sigma\left(\int_\Omega \kappa_\phi(x-y) \cdot z_t(y)dy + W \cdot z_t(x)\right), \quad t = 0,1,\ldots,L-1. \quad (5)$$

The convolution kernel function $\kappa_\phi(x-y)$ comes from the perspective of fundamental solutions of PDEs, and will be learned from data. If we parameterize $\kappa_\phi$ directly in Fourier space and using the Fast Fourier Transform (FFT) to efficiently compute the integral, we get the so called Fourier layer (see [17] Fig.2), and hence the name of FNO. Lastly, the hidden layer output $z_L$ is projected back to the target dimension of $u(x)$ by another neural network $Q$:

$$\mathcal{G}_\mathcal{N}(v(x)) = Q(z_L(x)). \quad (6)$$

### 4.2 Physics-guided Data Augmentation

Here we show how to augment the training data of neural operators when applied to learn the solution operator of linear differential equations. Different from image data augmenting techniques like flip, rotation, crop, etc, we augment the training data with prior physical information of PDEs like linearity and translation.

For a given PDE (1) defined on a bounded domain $\Omega$, we may assume $f=0$ by simple solution transformation techniques. The solution operator $\mathcal{G}$ with $u(x) = \mathcal{G}(v)(x)$ is linear since the operators $\mathcal{L}$ and $\mathcal{B}$ are both linear. Assume we have obtained a training set $\mathcal{T} = \{(v^{(1)}, u^{(1)}), (v^{(2)}, u^{(2)}), \cdots, (v^{(N)}, u^{(N)})\}$, we can then augment the training set by the linear combination of sampling solutions. The new training data can be expressed as

$$(c_1 v^{(i)} + c_2 v^{(j)}, c_1 u^{(i)} + c_2 u^{(j)}), \quad \forall c_1, c_2 \in \mathbb{R} \text{ and } i,j = 1,\ldots,N. \quad (7)$$

Another technique is to use the translation invariance of the PDE. We first obtain a sampled training data with constant inputs $v^{(0)} = 1$ on the domain $\Omega$, and the corresponding solution $u^{(0)} = \mathcal{G}(v^{(0)})$. Then we have the new training data

$$(v^{(i)} + v^{(0)}, u^{(i)} + u^{(0)}), \quad \forall i = 1,\ldots,N. \quad (8)$$

The augmentation technique (7) and (8) can be combined to produce more training data without additional cost to generate data

$$(c_0 v^{(0)} + c_1 v^{(i)} + c_2 v^{(j)}, c_0 u^{(0)} + c_1 u^{(i)} + c_2 u^{(j)}), \quad \forall c_0, c_1, c_2 \in \mathbb{R}; i,j = 1,\ldots,N \quad (9)$$

For the case when $\mathcal{L}$, $\mathcal{B}$ or the operator $\mathcal{G}$ are nonlinear, then the universal linearity and transformation augmentation techniques are no longer valid. The augmentations can be tailored to the physical properties of the PDE's solution. For example, for the Darcy flow problem $-\nabla \cdot (a(x)\nabla u(x)) = f(x)$ in porous media, we consider the nonlinear solution operator $\mathcal{G}: a(x) \to u(x)$, mapping from the permeability function $a(x)$ to the pressure function $u(x)$. If $(a(x), u(x))$ is a pair of training data, then we can augment it by $\left(c_0 a(x), \frac{1}{c_0} u(x)\right)$ with constant $c_0 \in \mathbb{R}, c_0 \neq 0$.

## 5 Experiments

We test our PGDA method on the solution operator learning of different linear differential equations, from a simple 1D linear ODE system to the 2D PDE dynamics and singular perturbation problems. We compare the testing error for neural operator models trained with/without PGDA and demonstrate that adding PGDA can improve generalization performance effectively.

**Evaluation Metrics.** The training and testing errors are measured by the following MSE error:

$$MSE = \frac{1}{N} \sum_{i=1}^{N} \left\| \mathcal{G}_\mathcal{N}(v^{(i)}) - u^{(i)} \right\|^2, \quad (10)$$

where $\|\cdot\|$ denotes the discrete $L^2$ norm in the output space, $v^{(i)}$ is the input function, $u^{(i)}$ is the target and $\mathcal{G}_\mathcal{N}(v^{(i)})$ is the DeepONet/FNO output.

**Experimental Setup.** We train DeepONet and FNO models with/without PGDA on different linear differential equations. All models are trained by Adam optimizer with an initial learning rate $\alpha = 0.001$ that is halved every 100 epochs. The training instances before augmentation are $N = 1,000$ and testing instances are 200. The DeepONet's trunk net and brunch net are 4 layers FNN with $p = 128$ and $m = 32$. The FNO has 4 Fourier layers with $k_{max} = 16$ and $d_v = 32$. The linear coefficients $c_0, c_1, c_2$ in the augmentation technique (9) are randomly sampled constants.

**Data Generalization.** The input function $v(x)$ is generated from the Gaussian Random Field (GRF)

$$v \sim \mathcal{G}(A, k_l(x_1, x_2)),$$

where the constant $A$ is the mean value of the sample function, and the covariance kernel $k_l(x_1, x_2) = \exp\left(-\frac{\|x_1 - x_2\|^2}{2l^2}\right)$ is the radial-basis function kernel with a length-scale parameter $l > 0$. We train DeepONet and FNO on one pair $(A_0, l_0)$ and test them both on the same pair $(A_0, l_0)$ and distributional shifted pairs $(A', l') \neq (A_0, l_0)$ with/without PGDA on different linear differential equations.

### 5.1 A Simple 1D Dynamic System

A 1D dynamic system is described by

$$\frac{du(x)}{dx} = g(u(x), v(x), x), \quad x \in [0,1], \quad (11)$$

with an initial condition $u(0) = 0$. Our goal is to predict $u(x)$ over the whole domain $[0,1]$ for any given $v(x)$. For simplicity we set $g(s(x), u(x), x) = v(x)$, then it is equivalent to learning the antiderivative operator $\mathcal{G}: v(x) \to u(x) = \int_0^x v(\tau)d\tau$.

We train DeepONet and FNO on the training data for $(A, l) = (0, 0.2)$ with and without PGDA, then we test the model on distribution shift data with different pairs $(A, l)$, see Table I. Both the DeepONet and FNO trained with PGDA can generalize very well on different test data, and the FNO with PGDA shows best performance.

### 5.2 Poisson Equation

The Poisson equation is an elliptic partial differential equation of broad utility in theoretical physics. We consider the Poisson equation on the domain $\Omega = [0,1] \times [0,1]$:

$$-\frac{\partial^2 u}{\partial x^2} - \frac{\partial^2 u}{\partial y^2} = f(x, y), \quad (x, y) \in \Omega, \quad (12)$$

with the Dirichlet boundary condition $u|_{\partial\Omega} = 0$. We want to learn the mapping $\mathcal{G}: f(x, y) \to u(x, y)$. Similar results as described in Section 5.1 can be obtained from Table I. An instance of the Poisson equation with a referenced solution and the learned neural operator solution are plotted in Figure 2.

**Table I** Test of PGDA on different tasks for improving generalization. MSE errors on different test sets are reported.

| Task | Test data | *Without* PGDA | | *With* PGDA | |
|---|---|---|---|---|---|
| | | DeepONet | FNO | DeepONet | FNO |
| Section 5.1 | $(A, l) = (0, 0.2)$ | $4.65 \times 10^{-4}$ | $1.68 \times 10^{-5}$ | $1.77 \times 10^{-3}$ | $\mathbf{9.44 \times 10^{-6}}$ |
| | $(A, l) = (0, 2)$ | $6.32 \times 10^{-4}$ | $1.45 \times 10^{-4}$ | $9.09 \times 10^{-4}$ | $\mathbf{1.27 \times 10^{-5}}$ |
| | $(A, l) = (10, 0.2)$ | $3.67 \times 10^{-3}$ | $3.85 \times 10^{-1}$ | $2.06 \times 10^{-4}$ | $\mathbf{3.33 \times 10^{-5}}$ |
| | $(A, l) = (10, 2)$ | $3.32 \times 10^{-3}$ | $3.86 \times 10^{-1}$ | $5.82 \times 10^{-5}$ | $\mathbf{1.42 \times 10^{-5}}$ |
| Section 5.2 | $(A, l) = (0, 0.2)$ | $1.52 \times 10^{-4}$ | $4.79 \times 10^{-6}$ | $1.75 \times 10^{-4}$ | $\mathbf{4.04 \times 10^{-6}}$ |
| | $(A, l) = (0, 2)$ | $4.19 \times 10^{-5}$ | $7.01 \times 10^{-6}$ | $4.03 \times 10^{-5}$ | $\mathbf{1.08 \times 10^{-6}}$ |
| | $(A, l) = (10, 0.2)$ | $3.23 \times 10^{-3}$ | $2.29 \times 10^{-1}$ | $1.03 \times 10^{-3}$ | $\mathbf{1.99 \times 10^{-5}}$ |
| | $(A, l) = (10, 2)$ | $3.03 \times 10^{-3}$ | $2.18 \times 10^{-1}$ | $9.05 \times 10^{-4}$ | $\mathbf{1.11 \times 10^{-5}}$ |
| Section 5.3 | $(A, l) = (0, 0.2)$ | $4.72 \times 10^{-3}$ | $4.01 \times 10^{-5}$ | $3.99 \times 10^{-3}$ | $\mathbf{1.21 \times 10^{-5}}$ |
| | $(A, l) = (0, 2)$ | $7.49 \times 10^{-3}$ | $2.81 \times 10^{-4}$ | $2.42 \times 10^{-3}$ | $\mathbf{1.68 \times 10^{-5}}$ |
| | $(A, l) = (10, 0.2)$ | $1.75 \times 10^{-1}$ | $4.29 \times 10^{-1}$ | $7.41 \times 10^{-3}$ | $\mathbf{4.76 \times 10^{-5}}$ |
| | $(A, l) = (10, 2)$ | $1.74 \times 10^{-1}$ | $4.28 \times 10^{-1}$ | $7.30 \times 10^{-5}$ | $\mathbf{5.07 \times 10^{-5}}$ |

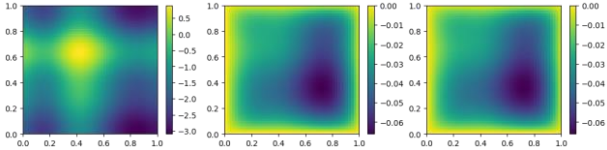

**Figure 2** The neural operator learning of the Poisson solution operator $\mathcal{G}: f(x, y) \mapsto u(x, y)$. (**Left**) An instance of input function $f(x, y)$. (**Middle**) Reference solution $(\mathcal{G}f)(x, y)$. (**Right**) Learned neural operator $(\mathcal{G}_\mathcal{N} f)(x, y)$ with PGDA.

### 5.3 Singularly Perturbed Advection-Diffusion Equation

Singularly perturbed problems have been successfully applied to many fields including gas dynamics, chemical reaction, fluid mechanics, elasticity, etc. To find the solution is a hot and difficult problem because it contains a very small parameter $0 < \epsilon \ll 1$. We consider the second-order linear singularly perturbed advection-diffusion equation

$$-\epsilon u''(x) + u'(x) = f(x), \quad x \in (0,1), \quad (13)$$

with Dirichlet boundary conditions $u(0) = u(1) = 0$. We want to learn the mapping $\mathcal{G}: f(x) \to u(x)$. As seen in Table I, MSE errors of DeepONet and FNO without PGDA increase for the appearance of this small parameter $\epsilon$, but our PGDA method still generalize well for different test data in the small parameter regime.

## 6  Conclusion and Future Work

In this paper we developed a physics-guided data augmentation (PGDA) technique for the neural operator model training in the application of differential equations to improve generalization. Training data is augmented based on the physical properties of differential equations such as linearity and translation. In the case with out-of-distribution test data, our PGDA methods generalize significantly better than models trained without data augmentation. Future work includes continuing to find other physical-properties for data augmentation and active sampling strategies to accelerate training process.

### Acknowledgements

This work is supported by the National Natural Science Foundation of China (No.62106103), Fundamental Research Funds for the Central Universities (No.ILA22023, No.90YAH20131), 173 Program Technical Field Fund (No.2021-JCJQ-JJ-0018).